# Cross-utterance Reranking Models with BERT and Graph Convolutional Networks for Conversational Speech Recognition

*Shih-Hsuan Chiu, Tien-Hong Lo, Fu-An Chao, and Berlin Chen*

Department of Computer Science and Information Engineering

National Taiwan Normal University, Taipei, Taiwan

{shchiu, teinhonglo, fuann, berlin}@ntnu.edu.tw

*Abstract*—How to effectively incorporate cross-utterance information cues into a neural language model (LM) has emerged as one of the intriguing issues for automatic speech recognition (ASR). Existing research efforts on improving contextualization of an LM typically regard previous utterances as a sequence of additional input and may fail to capture complex global structural dependencies among these utterances. In view of this, we in this paper seek to represent the historical context information of an utterance as graph-structured data so as to distill cross-utterances, global word interaction relationships. To this end, we apply a graph convolutional network (GCN) on the resulting graph to obtain the corresponding GCN embeddings of historical words. GCN has recently found its versatile applications in social-network analysis, text summarization, and among others due mainly to its ability of effectively capturing rich relational information among elements. However, GCN remains largely underexplored in the context of ASR, especially for dealing with conversational speech. In addition, we frame ASR *N*-best reranking as a prediction problem, leveraging bidirectional encoder representations from transformers (BERT) as the vehicle to not only seize the local intrinsic word regularity patterns inherent in a candidate hypothesis but also incorporate the cross-utterance, historical word interaction cues distilled by GCN for promoting performance. Extensive experiments conducted on the AMI benchmark dataset seem to confirm the pragmatic utility of our methods, in relation to some current top-of-the-line methods.

*Keywords*—automatic speech recognition, language modeling, *N*-best hypothesis reranking, cross-utterance, BERT, GCN

## I. INTRODUCTION

Dictation of conversational speech with automatic speech recognition (ASR) has many use cases in our daily lives. Possible applications range from meeting and dialogue transcriptions, interactive voice responses to smart speakers, just to name a few [1][2]. Successful deployment of these applications also predominantly hinges on the performance of ASR [3]. Meanwhile, a language model (LM) is an integral component of any ASR system since it can be employed to constrain the acoustic analysis, guide the search through multiple candidate word (or subword) strings, and quantify the acceptability of the final output from an ASR system [4]. The traditional *n*-gram LMs [5][6][7] are inadequate in modeling within-sentence (local) regularity patterns of language usage, since they determines the probability of a current word given its *n*-1 immediate predecessors. In this review, neural LMs instantiated with recurrent neural network (RNN) [8][9], long short-term memory (LSTM) [10][11][12], Transformer LM (TLM) [13][14], and others have aroused great attention recently to tackle this issue. These neural LMs, however, are hardly be used at the first-pass decoding stage of ASR due to the exponential growth of hypothesis search space with the increase in the modeling context of an LM. An alternative and lightweight therapy is to employ them to give LM scores to the hypotheses at the second-pass *N*-best hypothesis reranking stage [15][16][17], making most ASR modules remain unchanged while having a fast experiment turnover.

On a separate front, a more recent trend in dealing with conversational ASR problems is to infuse cross-utterance information cues into a neural language model when perform ASR on an utterance [18][19][20][21]. However, when with an RNN- or LSTM-based LM, context carryover of historical utterances by concatenating their LM hidden state representations or ASR transcripts for use in *N*-best hypothesis reranking often only leads to moderate improvements. This may because that RNN and LSTM inherently suffer from the sharp nearby, fuzzy far away issues [22] and may fail to capture complex global structural dependencies among these utterances. Building on these observations, we in this paper manage to render the historical context information of an utterance as graph-structured data so as to distill cross-utterances, global word interaction relationships. For the idea to work, a graph convolutional network (GCN) [23] is operated on the resulting graph to obtain the corresponding GCN embeddings of historical words. GCN has recently found its versatile applications on social-network analysis, text summarization, and many others due mainly to its ability of effectively extracting rich relational information among elements. To our knowledge, there is a dearth of work on tapping into GCN for language modeling of ASR, especially for handling conversational speech. Furthermore, ASR *N*-best reranking will boil down to a prediction problem in this study. To this end, we employ bidirectional encoder representations from transformers (BERT) [24][25][26] as the vehicle to not only seize the local intrinsic word regularity patterns inherent in a candidate hypothesis but also incorporate the cross-utterance, historical word interaction cues uncovered by GCN for better ASR performance. The rest of this paper is organized as follows. Section II introduces the BERT-based reranking framework, while Section III elucidates our proposed approach to capitalizing on GCN to model historical utterances for ASR *N*-best reranking. After that, the experimental setup and results are presented in Section IV. Finally, we summarize this paper and envisage future research directions in Section V.

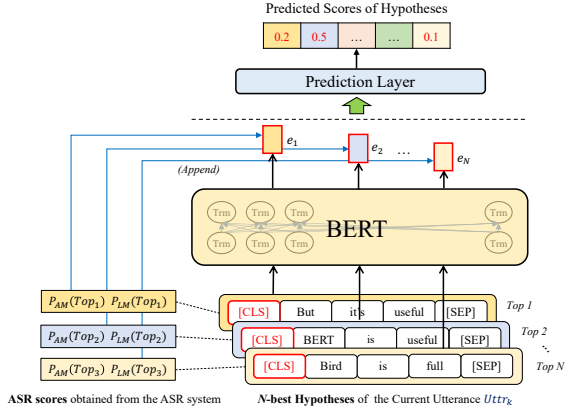

Figure 1: A schematic depiction of PBERT for ASR *N*-best hypothesis reranking [24].

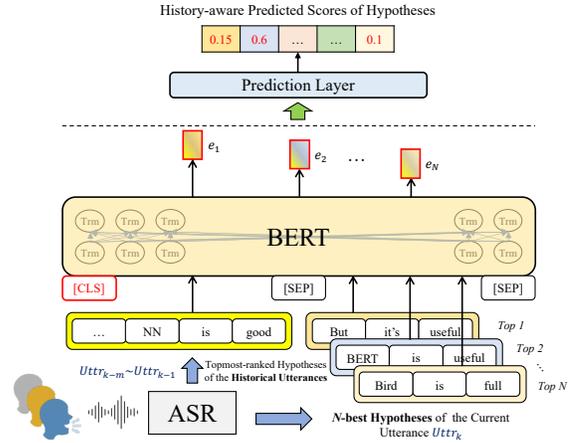

Figure 2: A schematic depiction of HPBERT for ASR *N*-best hypothesis reranking [25].

## II. BERT-BASED RERANKING FRAMEWORK

In this section, we review the fundamentals and instantiations of previously proposed BERT-based modeling framework for ASR *N*-best hypothesis reranking [24][25]. As we will see in Section III, we extend this framework by additionally incorporating the global information about the vocabulary and language structure inherent in the preceding utterances of an utterance of interest. Such global information is captured by GCN for better reranking.

### A. Fundamentals of BERT

BERT [26] is a neural contextualized language model, which leverages multi-layer Transformer encoder [27] based on the so-called multi-head self-attention mechanism. Such a mechanism is anticipated to simultaneously capture different aspects of local contextual interactions between lexical units (which are usually in the form of words or word pieces) involved in its input token sequence(s). In contrast to the traditional word embedding methods, the main advantage of BERT is that it can produce different contextualized representations of individual words at different locations by considering bidirectional dependencies of words within a sentence or across consecutive sentences. The training of BERT in general consists of two stages: pre-training and fine-tuning. At the pre-training stage, BERT is essentially trained to optimize the two tasks, namely the masked language modeling (MLM) and the next sentence prediction (NSP), on large-scale unlabeled text corpora (e.g., Wikipedia). At the fine-tuning stage, the pre-trained BERT model, stacked with an additional single- or multi-layer feedforward neural network (FFN), can be fine-tuned to work well on many text learning tasks when only a very limited amount of supervised task-specific training data is made available.

### B. BERT for ASR N-best Hypothesis Reranking

In recent work [24], an effective BERT-based modeling framework for ASR *N*-best hypothesis reranking had been put forward. This framework aims to predict a hypothesis that would have the lowest WER (i.e., the oracle hypothesis) from an *N*-best list (denoted by PBERT). In realization, PBERT consists of two model components, namely BERT stacked with an additional prediction layer which usually a simple fully-connected FFN, as depicted in Figure 1. For a given ASR *N*-best list, each hypothesis is respectively taken as an input to the BERT component, and meanwhile [CLS] and [SEP] tokens are inserted at the beginning and end of each hypothesis, respectively. In turn, the resulting embedding vector of [CLS] is used as a semantic-aggregated representation of the input hypothesis. After that, the [CLS] embedding vectors of all the *N*-best hypotheses are spliced together to be fed into the FFN component to output a prediction score (with the softmax normalization) for each hypothesis. Given a set of training utterances, each of which is equipped with an *N*-best hypothesis list generated by ASR and the indication of the oracle hypothesis that has the lowest WER, we can train the FFN component and fine-tune the BERT component accordingly.

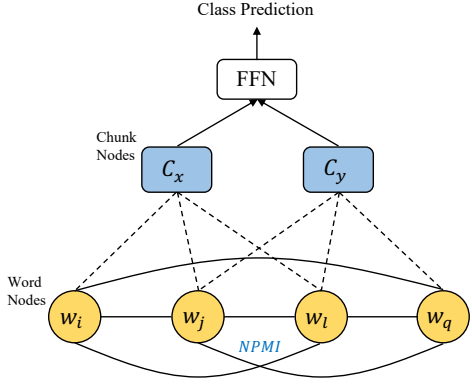

Figure 3: A schematic description of a heterogenous graph that contains chunk and word nodes.

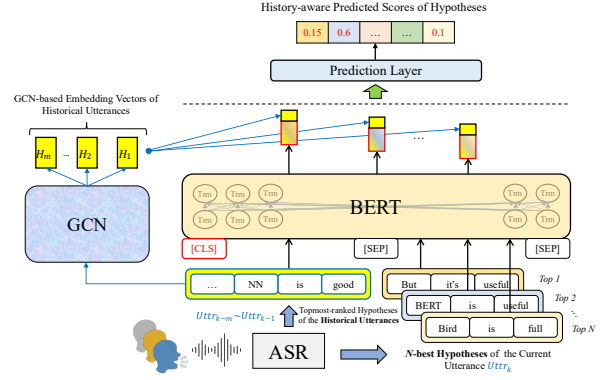

Figure 4: A schematic depiction of the incorporation of the GCN-based embedding vectors of the historical utterances as a whole into HPBERT for ASR *N*-best hypothesis reranking.

Not content to merely conducting ASR *N*-best hypothesis reranking for an utterance in isolation, the authors of [25] had taken a step further to incorporate historical, cross-utterance context information into PBERT (denoted by HPBERT), as schematically depicted in Figure 2. This is because that in real use cases of ASR, a sequence of consecutive utterances may jointly contain many important conversation-level phenomena across utterances. More specifically, each hypothesis of the current utterance of interest is first concatenated with the topmost ASR hypotheses (sequentially generated by *N*-best hypothesis reranking) of its preceding $m$ utterances with a special symbol [SEP] as the delimiter. Then, they as a whole are fed into the BERT component to generate a cross-utterance history-aware embedding (viz. [CLS] vector) of the hypothesis. Along this same vein, the resulting embeddings of all the *N*-best hypotheses are spliced together to be fed into the FFN component to output a prediction score for each hypothesis for the reranking purpose. Note also that the acoustic model score, language model score (i.e., *n*-gram or LSTMs) or their combination score for each hypothesis, obtained from ASR, can be concatenated together with the corresponding [CLS] vector of the hypothesis for feature augmentation.

## III. PROPOSED APPROACH

Although BERT has been proven powerful in capturing the contextual information within an utterance or consecutive utterances, their ability of capturing the global information about the vocabulary and structure of a language is relatively limited. In this review, we explore the use of graph embeddings of historical utterances generated by graph convolutional network (GCN) to augment the embedding vector of a hypothesis generated from PBERT or HPBERT (*cf.* Section II-B), with the goal to harness the synergistic power of BERT and GCN for ASR *N*-best hypothesis reranking.

### A. Fundamentals of GCN

Recently, there has been a surge of attempts in the literature to extend neural networks to deal with arbitrarily structured graphs [28]. One of the prevailing paradigms is the family of GCN [23][29][30]. GCN is instantiated with a multilayer neural network (usually consisting of 2 layers) that employs convolution operators on a target graph and iteratively aggregates the embeddings of the neighbors for every node on the graph to generate its own embedding. A bit of terminology: consider a graph $G = (V, E)$ that encompasses a set of nodes $V = \{v_1, v_2, ..., v_n\}$ and a set of edges $E = \{e_{i,j}\}$, where any given pair of nodes $v_i$ and $v_j$ is connected by an edge $e_{i,j}$ with a certain weight if they have a neighborhood relationship (or share some properties). We can represent the graph with either an adjacency matrix or a derived vector space representation. Furthermore, the (diagonal) degree matrix D of the graph G is defined by $D_{i,i} = \sum_j A_{i,j}$. For GCN equipped merely with a single-layer structure, the updated feature matrix of all nodes on G is calculated as follows:

$$H^{(1)} = ReLU(\tilde{A}XW_0), \quad (1)$$

where $X \in \mathbb{R}^{n \times m}$ is an input matrix that contains the corresponding *m*-dimensional feature vectors of all nodes in G, $W \in \mathbb{R}^{m \times k}$ is a weight matrix to be estimated, $\tilde{A} = D^{-\frac{1}{2}}AD^{-\frac{1}{2}}$ is the normalized symmetric adjacency matrix for G. The normalization operation that converts A to $\tilde{A}$ is to avoid numerical instabilities and exploding (or vanishing) gradients when estimating W of the corresponding GCN model in response to G. Building on this procedure, we can extend to capture higher-order neighborhood information from G by stacking multiple GCN layers:

$$H^{(i+1)} = ReLU(\tilde{A}H^{(i)}W_i), \quad (2)$$

where *i* denotes the layer number and $H^{(0)} = X$.

Table 1: Basic Statistics of the AMI Meeting Corpus.

| Items | Train | Dev | Eval |
|---|---|---|---|
| # Hours | 78 | 8.71 | 8.97 |
| # Utterances | 108,104 | 13,059 | 12,612 |
| # Speaker Turns | 10,492 | 1,197 | 1,166 |
| # Uttrs. Per (Spk. Turn) | 10.30 | 10.91 | 10.82 |
| # Words Per Utterance | 7.42 | 7.27 | 7.1 |

The authors of [23] first proposed semi-supervised learning of GCN for a node-level classification task. In addition, [29] regarded simultaneously documents and words of a text corpus as nodes to construct the corpus graph (a heterogeneous graph) and used GCN to learn embeddings of words and documents. It can capture global word co-occurrence information, given a limited number of labeled documents is provided.

*B. GCN-based Graph Embeddings for ASR N-best Hypothesis Reranking*

For its application to ASR *N*-best hypothesis reranking, the whole training process of GCN contains two primary stages. At the first stage, the entire LM training corpus is split into multiple non-overlapping chunks, each of which consists of a fixed number of consecutive sentences (e.g., 10 sentences). On top of the resulting chunks and the co-occurrence relationships among words in each chunk, we in turn construct a heterogeneous graph with chunk and word nodes. The co-occurrence relationship among any pair of word nodes $i$ and $j$ (with respect to a chunk or multiple chunks) is represented as an undirected edge with a weight that quantify their relatedness, which can be computed using a formula that is expressed by normalized point-wise mutual information (NPMI) [30][31]:

$$\text{NPMI}(w_i, w_j) = -\frac{1}{\log p(w_i, w_j)} \log \frac{p(w_i, w_j)}{p(w_i)p(w_j)}, \quad (3)$$

where $i$ and $j$ denote arbitrary two distinct words; $p(w_i) = \frac{\#W(i)}{\#W}$ ; $p(w_i, w_j) = \frac{\#W(i,j)}{\#W}$ ; $\#W(i)$ and $\#W(i,j)$ are the numbers of chunks respectively containing word $i$ and words $i$ and $j$; and $\#W$ is the total number of chunks. Note also that NPMI has its value that ranges from -1 to 1: the higher the value the closer the semantic relation between two words $i$ and $j$, and vice versa. On the other hand, an arbitrary chunk node $k$ has an edge connecting to a word node $i$ if the word $w_i$ is involved in the chunk $C_k$, where the weight of the edge is simply determined by the frequency that $w_i$ occurs in $C_k$.

At the second stage, the entire chunks of the training data are first clustered into a fixed number of groups using an off-the-shelf clustering algorithm (e.g., the *K*-means algorithm [32]) and TF-IDF vector representations for individual chunks. Each chunk is in turn assigned to a specific group, which is regarded as a "pseudo" class for the chunk. The model parameters of GCN are trained with the cross-entropy objective function, which aims to minimize the discrepancy between the reference "pseudo" class and the prediction output of a one-layer FFN module that takes the GCN-based embedding vector of every training chunk (*cf.* Section III-A) as the input to predict its own "pseudo" class, as graphically illustrated in Figure 3.

One the other hand, at test time (when performing ASR *N*-best hypothesis reranking), the corresponding embedding vector of each historical utterance is first obtained by a folding-in process that simply pools together the GCN-based embedding vectors of all words occurring in the topmost transcripts of the utterance. Following that, the GCN-based embedding vectors of all historical utterances as whole is a composition of the GCN-based embedding vectors of all historical utterances with an exponentially decayed weighting mechanism (the weight of the GCN-based embedding vector of an historical utterance is decayed exponentially with its distance from the historical utterance to the current sentence). Finally, the resulting GCN-based embedding vectors of all the historical utterances as whole is in turn appended to the BERT embedding of each hypothesis of the current sentence for the purpose of oracle hypothesis prediction. Figure 4 shows a graphical illustration of the incorporation of GCN-based embedding vectors of the historical utterances into HPBERT for ASR *N*-best hypothesis reranking.

## IV. EMPIRICAL EXPERIMENTS

*A. Experimental Setup*

We evaluate our proposed methods on the AMI benchmark meeting transcription database and task [33], while the baseline ASR system (which was employed to produce the *N*-best hypothesis list for any given utterance) was built with ASR toolkit Kaldi [34]. As to the AMI database, the speech corpus consisted of utterances collected with the individual headset microphones (IHM), while a pronunciation lexicon of 50K words was used. Table 1 shows some basic statistics of the AMI meeting corpus for our experiments.

The ASR system employed in this paper adopted the hybrid DNN-HMM modeling paradigm. For acoustic modeling, our recipe was to first estimate a GMM-HMM acoustic model on the training utterances with their corresponding orthographic transcripts, and in turn used the prior distribution of the senones obtained from the GMM components, in conjunction with the LF-MMI training objective, to estimate the factorized time-delay neural network (TDNN-F) acoustic model [35], which was taken as the seed model. The speech feature vectors were 40 MFCC coefficients extracted from frames of 25-msec length and 10-msec shift, spliced with 100-dimensional i-vectors for speaker adaptation of TDNN-F. On the other hand, the language model we used for the first-pass, baseline ASR system was a trigram LM, trained on the transcripts of AMI training utterances. The trigram LM was estimated with Kneser-Ney (KN) smoothing [5] and had a vocabulary of 12k words. To validate the improvements of our proposed LM modeling approach and have a fast experiment turnover, we adopt ASR

Table 2: The WER (%) results obtained by our proposed PBERT with GCN method for ASR *N*-best hypothesis reranking, in comparison to that of the LSTM- and BLSTM-based LM methods.

| Method | WER (%) |
| --- | --- |
| Baseline ASR | 18.67 |
| LSTM | 17.27 |
| BLSTM | 17.00 |
| PBERT | 16.48 |
| PBERT + GCN(1) | **16.26** |
| PBERT + GCN(5) | 16.29 |
| PBERT + GCN(10) | 16.27 |

Table 3: The WER (%) results obtained by HPBERT and HPBERT with GCN.

| Method | WER (%) |
| --- | --- |
| HPBERT(1) | 16.24 |
| HPBERT(5) | 16.16 |
| HPBERT(10) | 16.15 |
| HPBERT(1) + GCN(1) | 16.20 |
| HPBERT(5) + GCN(5) | 16.14 |
| HPBERT(10) + GCN(10) | **16.13** |

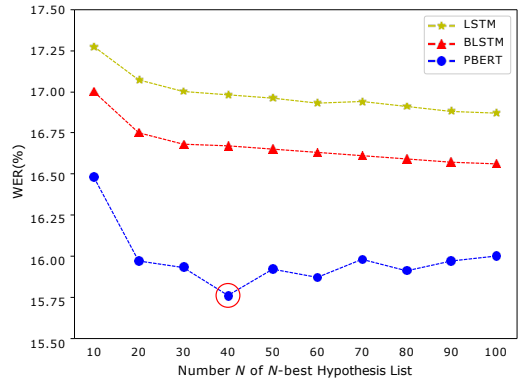

Figure 5: The performance levels of PBERT as a function of different numbers of ASR *N*-best hypotheses being considered, in comparison to that of two conventional autoregressive LM methods (i.e., LSTM and BLSTM).

*N*-best hypotheses with size *N*=10 for all reranking experiments. The *N*-best hypotheses for each utterance were extracted from the lattice generated by the ASR system during the test time.

*B. Experimental Results*

In the first set of experiments, we assess the efficacy of augmenting the GCN-based embedding vectors of the historical utterances as a whole to PBERT for ASR *N*-best hypothesis reranking (denoted by PBERT+GCN), in relation to some strong baselines that rerank *N*-best hypotheses with the conventional LSTM- and bidirectional-LSTM-based LM methods [11][12] (denoted by LSTM and BLSTM for short, respectively) and PBERT, respectively. In our default setting for PBERT (also the same for HPBERT), we augment the BERT output embedding of each hypothesis with a log-liner combination of the acoustic model score generated by the ASR system and the language model probability score generated by the trigram LM interpolated with the LSTM-based LM. The corresponding results are shown in Table 2, where the word error rate (WER) result of the baseline ASR system is listed for reference. For PBERT+GCN, we report on the results obtained by using different numbers of immediately preceding utterances for history modeling involved in GCN (*cf.* Rows 5 to 7). By looking at Table 2, we can make two noteworthy observations. First, PBERT offers a considerable improvement in terms of WER reduction over the conventional autoregressive LSTM- and BLSTM-based LM methods, which seems to confirm the feasibility of the PBERT-based reranking framework. Second, as historical sentences are made available, a substantial reduction of WER can be achieved by combing PBERT with GCN that makes additional use of different numbers of historical utterances. In particular, PBERT+GCN(1) stands out in performance, leading to a relative WER reduction of 1.3% over PBERT in isolation. This indeed demonstrates the efficacy of GCN-based embedding for historical sentences, which to some extent captures the global information about the vocabulary and language structure inherent in the historical utterances.

In the second set of experiments, we move on to evaluating the utility of augmenting the GCN-based embedding vectors of the historical utterances collectively to HPBERT for ASR *N*-best hypothesis reranking (denoted by HPBERT+GCN). The corresponding results are shown in Table 3, from which we can draw two important observations. First, when only one immediately preceding utterance is additionally made use for *N*-best hypothesis reranking, HPBERT(1) yields quite comparable performance to PBERT+GCN (*cf.* Row 5 of Table 2). Second, by augmenting the GCN-based embedding vectors of the historical utterances collectively to HPBERT, we can obtain slight but consistent improvements over HPBERT in isolation when different numbers of historical sentences are involved. This also reveals that information distilled from BERT and GCN is complementary to each other, and their combination is useful for ASR *N*-best hypothesis reranking.

As a side note in this section, we turn to study the performance levels of the BERT-based prediction framework (taking PBERT as an example) with regard to different numbers of ASR *N*-best hypotheses being considered, in comparison to that of two conventional autoregressive LM methods (i.e., LSTM and BLSTM). Consulting to Figure 5, we notice that the performance of LSTM and BLSTM is steadily improved when the number of ASR *N*-best hypotheses being considered for reranking become larger, while the relative improvements appear to diminish slowly. As to PBERT, it consistently outperforms the aforementioned two autoregressive neural LMs (viz. LSTM and BLSTM); however, its relative improvement reaches the maximum when the *N* is set to 40.

## V. CONCLUSION AND FUTURE WORK

In this paper, we have presented a novel and effective modeling approach to ASR *N*-best hypothesis reranking, which leverages GCN-based embeddings to represent historical utterances. GCN holds promise to capture the global information about the vocabulary and language structure inherent in the historical utterances. Experimental results on a benchmark meeting transcription task indeed show the practical utility of our proposed approach in comparison to some top-of-the-line LM methods. In future work, we envisage to explore more sophisticated techniques for better representation of contextual information that resides in a cross-utterance history for ASR.

## VI. ACKNOWLEDGEMENT

This research is supported in part by Chunghwa Telecom Laboratories under Grant Number TL-110-D301, and by the National Science Council, Taiwan under Grant Number MOST 109-2634-F-008-006- through Pervasive Artificial Intelligence Research (PAIR) Labs, Taiwan, and Grant Numbers MOST 108-2221-E-003-005-MY3 and MOST 109-2221-E-003-020-MY3. Any findings and implications in the paper do not necessarily reflect those of the sponsors.

## REFERENCES

[1] G. Saon, G. Kurata, T. Sercu, K. Audhkhasi, S. Thomas, D. Dimitriadis, X. Cui, B. Ramabhadran, M. Picheny, L.-L. Lim, B. Roomi, and P. Hall, "English conversational telephone speech recognition by humans and machines", in Proceedings of the Annual Conference of the International Speech Communication Association (Interspeech), 2017.

[2] W. Xiong, J. Droppo, X. Huang, F. Seide, M. Seltzer, A. Stolcke, D. Yu, and G. Zweig, "Achieving human parity in conversational speech recognition," IEEE/ACM Transactions on Audio, Speech, and Language Processing (TASLP), 2016.

[3] A. Becerra, J. I. de la Rosa, and E. González, "Speech recognition in a dialog system: From conventional to deep processing," Multimedia Tools and Applications, vol. 77, no. 12, pp. 15875–15911, Jul. 2018.

[4] B. Chen and J.-W. Liu, "Discriminative language modeling for speech recognition with relevance information," in Proceedings of the IEEE International Conference on Multimedia & Expo (ICME), Barcelona, Spain, 2011.

[5] R. Kneser and H. Ney, "Improved backing-off for m-gram language modeling," in Proceedings of the IEEE International Conference on Acoustics, Speech and Signal Processing (ICASSP), pp. 181-184, 1995.

[6] S. F. Chen and J. Goodman, "An empirical study of smoothing techniques for language modeling," Computer Speech & Language, vol. 13, no. 4, pp. 359–394, 1999.

[7] J. T. Goodman, "A bit of progress in language modeling," Computer Speech & Language, vol. 15, no. 4, pp. 403–434, 2001.

[8] T. Mikolov, M. Karafiát, L. Burget, J. Černocký, and S. Khudanpur, "Recurrent neural network based language model," in Proceedings of the Annual Conference of the International Speech Communication Association (Interspeech), 2010.

[9] T. Mikolov, S. Kombrink, L. Burget, J. Černocký, and S. Khudanpur, "Extensions of recurrent neural network language model," in Proceedings of the IEEE International Conference on Acoustics, Speech and Signal Processing (ICASSP), pp. 5528–5531, 2011.

[10] S. Hochreiter and J. Schmidhuber, "Long short-term memory," Neural Computation, vol. 9, no. 8, pp. 1735–1780, 1997.

[11] M. Sundermeyer, R. Schluter, and H. Ney, "LSTM neural networks for language modeling," in Proceedings of the Annual Conference of the International Speech Communication Association (Interspeech), 2012.

[12] E. Arisoy, A. Sethy, B. Ramabhadran, and S. Chen, "Bidirectional recurrent neural network language models for automatic speech recognition," in Proceedings of the IEEE International Conference on Acoustics, Speech and Signal Processing (ICASSP), pp. 5421–5425, 2015.

[13] K. Irie, A. Zeyer, R. Schlüter, and H. Ney, "Language modeling with deep transformers," in Proceedings of the Annual Conference of the International Speech Communication Association (Interspeech), 2019.

[14] K. Li, Z, Liu, T. He, H. Huang, F. Peng, D. Povey, and S. Khudanpur, "An empirical study of transformer-based neural language model adaptation" in Proceedings of the IEEE International Conference on Acoustics, Speech and Signal Processing (ICASSP), 2020.

[15] L. Liu, Y. Gu, A. Gourav, A. Gandhe, S. Kalmane, D. Filimonov, A. Rastrow, and I. Bulyko, "Domain-aware neural language models for speech recognition," in Proceedings of the IEEE International Conference on Acoustics, Speech and Signal Processing (ICASSP), 2021.

[16] A. Shenoy, S. Bodapati, M. Sunkara, S. Ronanki, and K. Kirchhoff, "Adapting long context NLM for ASR rescoring in conversational agents," in Proceedings of the Annual Conference of the International Speech Communication Association (Interspeech), 2021.

[17] A. Shenoy, S. Bodapati, and K. Kirchhoff, "ASR adaptation for E-commerce chatbots using cross-utterance context and multi-task language modeling," in Proceedings of the Annual Meeting of the Association for Computational Linguistics (ACL), 2021.

[18] W. Xiong, L. Wu, J. Zhang, and A. Stolcke, "Session-level language modeling for conversational speech," in Proceedings of the Conference on Empirical Methods in Natural Language Processing (EMNLP), 2018.

[19] K. Irie, A. Zeyer, R. Schlüter, and H. Ney, "Training language models for long-span cross-sentence evaluation," in Proceedings of the IEEE workshop on Automatic Speech Recognition and Understanding (ASRU), 2019.

[20] S. Kim, S. Dalmia, and F. Metze, "Gated embeddings in end-to-end speech recognition for conversational-context fusion," in Proceedings of the Annual Meeting of the Association for Computational Linguistics (ACL), 2019.

[21] S. Kim, S. Dalmia, and F. Metze, "Cross-attention end-to-end ASR for two-party conversations," in Proceedings of the Annual Conference of the International Speech Communication Association (Interspeech), 2020.

[22] U. Khandelwal, H. He, P. Qi, D. Jurafsky, "Sharp nearby, fuzzy far away: How neural language models use context," in

Proceedings of the Annual Meeting of the Association for Computational Linguistics (ACL), 2018.
[23] T. N. Kipf and M. Welling, "Semi-supervised classification with graph convolutional networks," in Proceedings of International Conference on Learning Representations (ICLR), 2017.
[24] S.-H. Chiu and B. Chen, "Innovative BERT-based reranking language models for speech recognition," in Proceedings of the IEEE Spoken Language Technology Workshop (SLT), 2021.
[25] S.-H. Chiu, T.-H. Lo, and B. Chen, "Cross-sentence neural language models for conversational speech recognition," in Proceedings of the IEEE International Joint Conference on Neural Networks (IJCNN), 2021.
[26] J. Devlin, M.-W. Chang, K. Lee, and K. Toutanova. "BERT: Pre-training of deep bidirectional transformers for language understanding," in NAACL-HLT, 2019.
[27] A. Vaswani et al., "Attention is all you need," in Proceedings of the International Conference on Neural Information Processing Systems (NIPS), 2017.
[28] J. Zhou et al., "Graph neural networks: A review of methods and applications," AI Open, vol. 1, pp. 57–81, 2021.
[29] L. Yao, C. Mao, and Y. Luo, "Graph convolutional networks for text classification," in Proceedings of the AAAI Conference on Artificial Intelligence (AAAI), 2019.
[30] Z. Lu, P. Du, and J.-Y. Nie, "VGCN-BERT: Augmenting BERT with graph embedding for text classification," In Advances in Information Retrieval - 42nd European Conference on IR Research (ECIR), 2020.
[31] G. Bouma, "Normalized (Pointwise) mutual information in collocation extraction," in Proceedings of the Biennial GSCL Conference, 2009.
[32] J. MacQueen, "Some methods for classification and analysis of multivariate observations," in Proceedings of 5th Berkeley Symposium on Mathematical Statistics and Probability, vol. 1, pp. 281-297, 1967.
[33] J. Carletta, S. Ashby, and S. Bourban et al., "The AMI meeting corpus: A pre-announcement," in Proceedings of the International Workshop on Machine Learning for Multimodal Interaction, pp. 28–39, 2005.
[34] D. Povey, A. Ghoshal, G. Boulianne, L. Burget, O. Glembek, N. Goel, M. Hannemann, P. Motlicek, Y. Qian, P. Schwarz, J. Silovsky, G. Stemmer, and K. Vesely, "The Kaldi speech recognition toolkit," in Proceedings of the IEEE Workshop on Automatic Speech Recognition and Understanding (ASRU), 2011.
[35] D. Povey, G. Cheng, Y. Wang, K. Li, H. Xu, M. Yarmohamadi, and S. Khudanpur, "Semi-orthogonal low-rank matrix factorization for deep neural networks," in Proceedings of the Annual Conference of the International Speech Communication Association (Interspeech), 2018.